# A Study Investigating Typical Concepts and Guidelines for Ontology Building

**Thabet Slimani**

[1]Asstt Prof., Department of Computer Science, College of Computer Science and Information Technology, Taif University

E-mail: thabet.slimani@gmail.com

## ABSTRACT

In semantic technologies, the shared common understanding of the structure of information among artifacts (people or software agents) can be realized by building an ontology. To do this, it is imperative for an ontology builder to answer several questions: a) what are the main components of an ontology? b) How an ontology look likes and how it works? c) Verify if it is required to consider reusing existing ontologies or no? c) What is the complexity of the ontology to be developed? d) What are the principles of ontology design and development? and e) How to evaluate an ontology? This paper answers all the key questions above. The aim of this paper is to present a set of guiding principles to help ontology developers and also inexperienced users to answer such questions.

## 1. INTRODUCTION

In computer science and information science, an ontology is defined as a structured set of terms and concepts representing the meaning of information domains, whether by the metadata of a namespace, or the elements of a knowledge domain. Ontologies are used in artificial intelligence, semantic web, software engineering, biomedical computing and information architecture as a form of knowledge representation about a world or a certain part of this world. The original definition of ontology is given by Gruber (1993) as an explicit specification of a conceptualization [1]. They offer a common vocabulary of a real world domain and define the meaning of the terms and the relations between them with a different degree of formality. Ontologies are usually organized in taxonomies and include the primitives of modeling such as classes, relations, functions, axioms and instances [1]. Knowledge management, knowledge-based systems, ontology-based brokers, and interoperability between systems are good examples of popular applications of ontologies.

At least, an ontology is a hierarchy of concepts description related by subsumption relationships; in more complicated models, appropriate axioms are added in order to make easy other relationships between concepts. Although the use of ontologies suggests a concrete approach to build sharable knowledge bases, it also raises a number of questions:
- What is the origin of an ontology?
- How should it be designed?
- What tools should be used for its construction?
- What design principles should I follow to build ontologies?
- How an ontology should be used?
- How should it be developed and evaluated?

The rest of this paper is organized as follows: In the next section, we describe the theoretical aspect of ontologies, their definitions, their reuse, their typologies and their complexities. Then ontology design principles are presented. In the following section, ontology development is described. After that, ontology evaluation principles are explained. And finally, this paper is wrapped up with a conclusion.

## 2. ONTOLOGY: THEORETICAL BASIS

### 2.1. Definitions

In the context of Artificial Intelligence, an ontology is identified with a set of formal terms representing knowledge. Gruber [1], defines an ontology as an explicit specification of a conceptualization. A conceptualization is an abstract view of the world that we hope to represent for some purpose. The knowledge representation is based on a conceptualization including objects, concepts, and the relationships that hold between them. Based on technical perspective, the definition of Sowa [2] considers an ontology as a specification of the kinds of entities that exist or may exist in some domain or subject area. More formally, an ontology can be represented by a collection of names designing concepts and relation types controlled in a partial ordering by the type/subtype relation. We will summarize some selected definitions of the well-known ontology definitions from many papers and discussions in academia.

**Definition1:** According Guarino [3] an ontology is defined as an engineering artifact, constituted by a specific vocabulary used for describing certain reality, in addition to a set of explicit assumptions regarding the intended meaning of the vocabulary words.

**Definition2:** According Neches et al. (1991) [4], ontologies are considered as a kind of "top-level declarative abstraction hierarchies represented with sufficient information to lay down the ground rules for modeling a domain". However, an ontology defines the basic terms and relations, including the vocabulary of a topic area as well as the rules for combining terms and relations to define extensions to the vocabulary.

**Definition3:** Wielinga and Schreiber (1993) [5], defined AI ontology as "a theory of what entities can exist in the mind of a knowledgeable agent". An ontology can be referenced as a meta-model describing the structure of a knowledge base and making explicit the commitments used in the modeling process which enables knowledge reuse and sharing.

**Definition4:** the definition of Gruber's [1] was redefined by Guarino and Giaretta (1995) [6]. They tried to refine the meaning of ontology, taking in consideration 7 senses of the terms that possibly to be used in the literature: Ontology
- as a philosophical discipline;
- as an informal conceptual system;
- as a formal semantic account;
- as a specification of a conceptualization;
- as a representation of a conceptual system via logical theory;
- as the vocabulary used by a logical theory;
- as a specification of a logical theory.

The authors' use directly the senses from 2 to 7 with an additional attention accorded to the sense 4 which they find problematic. Therefore, Guarino and Giaretta describe AI ontology, as having two senses:
- Ontology is a logical theory represented as a "designed artifact, a knowledge base of a special kind which can be read, sold or physically shared.".
- Ontology is a synonym of conceptualization defined as "an intentional semantic structure which encodes the implicit rules constraining the structure of the piece of reality."

Yet, according to Guarino and Giaretta, the same ontological theory may belong to different conceptualizations and the same conceptualization may underlie different ontological theories.

## 2.2. Ontology components

The formalization of Knowledge in ontologies is based on five components: classes, instances, relations, functions, axioms and instances [1]. The following paragraphs enumerate those ontology components:
- **Classes or concepts:** also called types or universals, are a group of individuals that share common characteristics used in a wide sense. A concept can be anything about which something is said. It can be a description of a task, action, function, strategy, reasoning process, etc. The majority of ontology languages (for example, OWL, DAML, etc.) allow the definition of concepts on the basis of these characteristics. For example, all mammals share the same characteristics, except for the ability to talk.
- **Relations:** In Ontologies, relations describe the means in which individuals (instances or particulars) are related. In other terms, relations represent a form of interaction between concepts in the same domain. Formally, a relation is any subset of a product of n sets, defined as follows: R: C1 x C2 x ... x Cn. Several types of relations can be expressed on Ontologies: "*subclass-of*" and "*connected-to*" are two examples of binary relations. The sentence "Ahmed teaches AI Course" express a direct relation between individuals, but the sentence "Professor teach Course" express a relation between Concepts.
- **Functions**: are a particular type of relations, where the $n^{th}$ element of the relationship is distinctive for the n-1 preceding elements. The relations "Author-of" and "Price-of-a-new-computer" are two examples of functions. For example, the second function aims to calculate the price of a new computer depending on the CPU type and speed, hard-disk storage capacity and the capacity of memory storage.
- **Axioms:** Axioms represent assertions formulated in a logical form that together comprise the core knowledge that the ontology describes in its domain of application. In other terms, axioms are used to model sentences that are always true. Axiom types can be classified according to their semantic meaning [7].
- **Instances:** instances are individuals that models concrete objects (people, proteins, machines) and represents the base components of an ontology. If the main components of ontologies have been represented, then it is possible implementing an ontology in various languages: highly informal, semi-informal, semi-formal and rigorously formal languages [8].

## 2.3. Ontology reuse

Ontology reuse is an important research issue in the semantic web and ontology field. It can be defined as the process in which available (ontological) knowledge is employed as input to generate new ontologies. In the following paragraphs, some definitions of ontology reuse categories are given (based on the work described in [9]):

- **Ontology merging**: It is an approach aiming to integrate two ontologies or more to create a unique ontology which would have/included all the knowledge that the merged ontologies had. Then, a process of ontology merging builds a unique ontology represents a merged version of the original ontologies. The obtained ontology contains all the information from merged original ontologies, without indication of their former origin [10]. A merging process is usually performed when the original ontologies cover similar or overlapping domains.
- **Ontology integration:** Ontology integration is a process creating a unique ontology after aggregation, assemblage, extension, combination, specialization or adaptation of ontologies on different subjects [10].
- **Ontology mapping**: Ontology mapping (OM) is a process identifying the correspondence among ontologies entities. Ontology mapping uses multiple ontologies to do inter-ontology mappings, but without the existence of a global ontology. As an output, the process of OM produces a set of mapping assertions denoting relations between these entities [11]. However, the aim of OM is to allow ontologies sharing, exchanging and reusing information between them.
- **Ontology alignment/matching:** ontology alignment is a process of determining correspondences between concepts contained in two ontologies aiming to modify one of them to make it more consistent and coherent with the other one [12]. Several tools of ontology alignment have been developed to operate on database schemas, XML schemas, taxonomies, entity-relationship models, formal languages, dictionaries, and other label frameworks.
- **Ontology versioning**: ontology versioning indicates that there is multiple variant of an ontology around. In reality, those variants often originate from changes to an existing variant of the ontology and thus form a derivation tree. However, ontology versioning is a process handling change in different versions of an ontology, which implies versions recognition, ontologies update and versions relationships traceability [13].

## 2.4. Ontology typologies

In this paper, the concept of ontological commitment is used to classify ontologies. If it is possible to differentiate between types of ontological commitment, then it is possible to identify different types of ontologies, and therefore, it is possible to classify ontologies. To define the different types of commitments, the commonly made distinction in AI literature between tasks, methods, and domains is adopted [5][14] (see Fig.1). Consequently, three different types of ontological commitments: task commitments, method commitments, and domain commitments can be identified. In addition to the three ontological commitments, this section will present a selected most commonly used types of ontologies as presented in [15].

- **Task commitments:** If an ontology defines the entities and relations expressing a task-specific perspective on the domain knowledge, then it has tasked commitments [16]. A task can be defined as a specification of a goal, including some input and required output. An ontology for a diagnosis task, including entities such as observations, hypotheses and causes is an example of an ontology that has task commitments.

- **Method commitments**: If an ontology defines the entities and relations expressing a method-specific perspective on the domain knowledge, then it has method commitment [16]. A method can be specified as a specification of how a task can be performed. An ontology adopting the propose and-revise method (within a design task), which contains entities such as proposed solution, value-assessment and constraints is an example of ontology that has method commitments.

- **Domain commitments**. : If an ontology defines entities and relations relating to a particular domain, then it has domain commitments. A domain is referred to the commonly distinguished fragments of the real world modeled, such as medical, judiciary, financial, mathematical or social domains. An ontology for the judiciary domain, which contains entities such as norms and acts, is an example of ontology that has domain commitments. Additionally, domain ontology supply vocabularies about the concepts of a given domain and their relationships, about the activities that take place in that domain, and about the theories and basic principles that governs the domain.

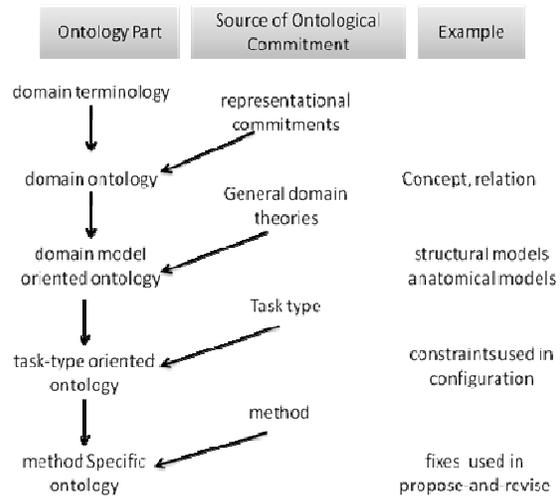

Fig1. Ontologies and ontological commitments by [5].

Based on the typology presented above and interesting to the development of KBS, Mizoguchi et al. (1995) [17] proposed a typology of ontologies based on the following idea: "from knowledge base technology perspective, knowledge should be considered in some context, that is, in problem solving situation". From these perspective, the following typologies are established:

1. Content ontology
   Task ontology
      Generic noun
      Generic verb
      Generic adjective
      Others
   Workplace ontology[22]
      Domain ontology
      Object ontology
      Activity ontology
      Field ontology
   General/Common ontology
      Things
      Events
      Time
      Space
      Causality
      Behavior
      Function
      etc.
2. Tell&Ask ontology
3. Indexing ontology
4. Meta-ontology

Mizoguchi et al. (1995)'s Ontologies typology.

- **Content Ontologies**: Also called ontologies for knowledge reuse, they include as well task ontologies, workplace (domain ontology) and general/common Ontologies (including vocabulary related to things, events, space, time, behavior, causality function, etc.).

The authors in [17] also take into account the concept of meta-ontology (also called Generic Ontologies or Core Ontologies) as a fourth main type to refer to representation Ontologies.

- **Communication Ontologies:** Also called knowledge sharing Ontologies. This type of ontologies is intended to support the sharing and reuse of represented knowledge in a formal manner. They are valuable to define the common vocabulary in which shared knowledge is represented. Ontolingua [18] is an example of Ontologies that adopts knowledge sharing.
- **Indexing Ontologies:** also called ontologies for information retrieval. The aim of adopting ontologies in Information Retrieval is to improve recall and precision [19,20]. Its major use is related to query expansion, consisting to look for the terms in the ontology more related to the query terms, and therefore using them as a part of the query.

### 2.5. Ontology complexity

Two types of ontology complexity can be distinguished depending on the broad range of tasks to which the Ontologies are included. The complexity range of Ontologies varies from simple taxonomies to highly complexed including constraints associated with concepts and relations.

- **Light-weight Ontologies:** Light-weight Ontologies are normally defined as more hierarchical or classificatory characteristic. They are habitually designed to represent subsumption relation (is-a hierarchy between concepts) or other types of relationships between concepts. A light-weight Ontology does not include too many or too complicated relationships. As example, those Ontologies can be applied in web search engines like Yahoo ontology consisting of a hierarchy of topics with little consideration of rigorous concept definition, distinction between word and concept, etc. A formal definition of lightweight ontology is given in [21]:
- **Heavy-weight Ontology:** The second type of Ontologies is different. Heavy weight ontology is developed with an additional attention to the rigorous meaning, organizing principles of each concept and semantically rigorous relations between concepts (cardinality constraints, taxonomy of relations, Axioms or restrictions, etc.). Target world modeling requiring a world conceptualization of the world to guarantee the consistency and fidelity of the model, needs an instance models typically built on heavy-weight ontology. In other terms, as relationships are added and the complexities of the world are increased, Ontologies migrate from the lightweight to the "heavyweight" complexity.

## 3. ONTOLOGY DESIGN

Ontologies can be defined as artifacts that have a structure (logical, linguistic, "taxonomical"). Their purpose is to encode a description of the world (actual, counterfactual, possible, impossible, desired, etc.) for some tasks or problems (e.g. the world of Medicine, the world of Semantic Web Conference, etc.). However, ontologies must match both domain and task described as follows:

- **Domain:** allowing the description of the entities whose attributes and relations are concerned by some purpose (student as entities that are enrolled in university, supervised by academic staff, and has name, address, etc.).
- **Task:** Serving a purpose (find persons working on a same topic, matching project topics to staff competencies, etc).

Ontology design is a quite challenge for complex applications and essential technique in the creation of knowledge-based applications [22]. Several area of research have been successfully applied Ontology design, such as model checking and semantic reasoning [23] and inconsistent detection in complex scene modeling [24][25]. The following section summarizes some design criteria and principles that have been proved useful in the development of Ontologies:

- **Clarity and Objectivity**: Based on the presentation in [26], Clarity signifies that the ontology should give the meaning of defined terms by objective definitions providing and also natural language documentation.
- **Completeness** [26]: Completeness means that a the expression of a definition is based on necessary and sufficient conditions that is preferred over a partial definition.
- **Coherence** [26]: Coherence is adopted to permit inferences that present consistence with the definitions.
- **Maximum monotonic extendibility** [26]: This property means that new general or specialized terms should be incorporated in the ontology in a fashion that is does not require the change of existing definitions.
- **Minimal ontological commitments** [26]: which means to minimize the claims about the world being modeled, giving the parties devoted to the ontology freedom specializing and instantiating the ontology as required.
- **Diversification of hierarchies** [27]: Diversification of hierarchies is useful to increase the power supplied by multiple inheritance mechanisms.
- **Principle of Ontological Distinction** [28]: this principle means that classes in ontologies should be disjoint.
- **Minimization of the semantic distance between sibling concepts** [27]: This principle means that the same primitives are grouped and used to represent similar concepts.
- **Modularity** [29] : This property is employed to minimize coupling between modules.
- **Standardization of names** [27].

## 4. ONTOLOGY DEVELOPMENT

A great number of methods and practices have been developed for ontology engineering in literature. We have selected few well known methods for explanation. The important thing is that all of these methods require that Ontologies are created either by a skilled ontology engineers, or by the cooperation between ontology engineers and domain experts. Only some of the methods support domain experts. Requirements engineering plays an important role in all of the presented methods. Gruninger and Fox [30] have introduced competency questions as a set of problems that the ontology logic axioms should be able to represent and solve. Based on the perspectives of Gruninger and Fox, such question must be formalized into machine interpretable and solvable problems, rather than natural language format. Competency questions in RDFS and OWL Ontologies are often formalized into SPARQL queries, and if SPARQL query returns the expected result after execution over the ontology in question they are considered satisfied.

Noy & McGuinness (2001) [31] address basis for developing Ontologies and enumerate the steps implicated in developing ontology. Based on that basis, the following section provides seven steps for ontology development described as follows:

- **Determine the domain and scope of the Ontology:** this step should determine the domain that the ontology will cover, specify the types of questions that the information in the ontology should provide answers (ontology useful for testing), specify for whom this ontology is intended (Users could be restaurant customers, professionals (chefs)) and verify if the ontology enclose enough information to answer these types of questions.
- **Consider the reuse of existing Ontologies:** This step should identify if the concerned system needs to interact with other applications that have already devoted to particular Ontologies or restricted vocabularies (Ontolingua library, DAML library, etc.). Moreover, this step verifies if exist a requirement to refine and extend existing sources for a particular domain.
- **Enumerate key terms in the ontology:** This step should identify a list of terms that would like to be explained to a user describing statements about a specific subject. Moreover, it should specify the properties enclosed in the terms and their interpretation (diverse types of food, such as fish and red meat). The list of terms is identified independently to the class categorization (hierarchy, facet and overlapping).
- **Classes and class hierarchy definition**: This step has to choose between three possible approaches: top-down development process which starts with the most general concept definition, bottom-up development process which starts with the most specific concept definition, or a combined approach between the first two approaches.
- **Class properties definition**: This step starts by the definition of classes from the list created in the step 3 and their internal structure (properties). A class is defined by describing objects that have independent existence. Moreover, in this step requires organizing classes into hierarchical structure (Taxonomy). A slot is a word which is not classes, and furthermore it should be assigned to a class.
- **Define the facet of slots**: Different facets can characterize a slot: the value type description (the types of values permitted to be filled in the slot: Number (e.g. for price)), permitted values, cardinality (defines the values number a slot can have), and other additional features of the values the slot can take (i.e. the value of a name is a string).
- **Create instances**: This step defines an individual instance of a class based on the following order: choose a class, then create an individual instance of that class, and finally fill it in the slot values.

## 5. ONTOLOGY EVALUATION

The evaluation process of am ontology could be based on different viewpoints: for example, the quality of the designed ontology (technical point of view) and its usability in real world (practical view). Additionally, it is possible to evaluate an ontology based on mathematical model (mathematical point of view). Several contributions have been proposed for the purpose of evaluation of quality of the designed ontology: For example, five criteria suggested by Gruber [26] needed for ontology evaluation: clarity, coherence, extendibility, minimal encoding bias and minimal ontological commitment. Ontologies include sets of related concepts. Generally, Set Theory is used to construct mathematical models to defining and reasoning about ontological models [32]. Moreover, Tartir et al. (2005) [33], presents OntoQA, as an approach designed for the purpose of quality evaluation of Ontologies and their potential for knowledge representation through a well defined set of metrics. The proposed approach divides the metrics into three categories:

- ❖ **Schema metrics:** addressing the design of ontology and composed by metrics that indicate the relationship richness, attribute richness and inheritance richness.
- ❖ **Knowledge base metrics**: describe the knowledge base as a whole and composed by class richness, average population and cohesion metrics.
- ❖ **Class metrics:** Also called instance distribution since it refers to the distribution of instances over classes. It includes metrics like fullness, inheritance richness, relationship richness, connectivity and readability.

Another well-known ontology evaluation approach presented in OntoClean [34] is inspired from a philosophical approach that evaluates formal properties of taxonomy. OntoClean, from a useful perspective, offers means to derive measurable mismatches of a taxonomy taking into account the semantics provided by the "is-a" relationship. Moreover, OntoClean offers a clarification of why mismatches occurs which subsequently improve the taxonomical structure. OntoClean properties are adopted by some tools like the works in[35], OntoEdit [36] and WebODE [37].

## 6. CONCLUSION

Ontologies have been exploited in several real world applications to help the communication improvement between agents (people or software agents). However, in literature several kinds of ontologies have been investigated and evolved in incremental manner. This paper was presented the concept of ontologies and their components, reuse, typology and complexity. Moreover, it describes the principle of ontology design and development guidelines. In addition, this paper has provided the process of ontology evaluation to help the understanding of ontology developer and naïf user to improve his understanding.

tional Conference on Knowledge Capture (K-CAP) Oct. 21–23, 2001, Victoria, B.C., Canada, 2001.

**Thabet Slimani.** got a PhD in Computer Science (2011) from the University of Tunisia. He is currently an Assistant Professor of Information Technology at the Department of Computer Science of Taif University at Saudia Arabia, where he is involved both in research and teaching activities. His research interests are mainly related to Semantic Web, Data Mining, Business Intelligence, Knowledge Management, Web services and Ontologies. Thabet has published his research through international conferences, chapter in books and peer reviewed journals. He also serves as a reviewer for some conferences and journals.